\documentclass[runningheads]{llncs}
\usepackage{graphicx}
\usepackage{comment}
\usepackage{amsmath,amssymb} 
\usepackage{color}
\usepackage{subfigure}
\usepackage{hyperref}
\usepackage{array}
\usepackage{epsfig}
\usepackage{epstopdf}
\usepackage{marvosym}

\begin{document}
\pagestyle{headings}
\mainmatter
\def\ECCVSubNumber{4600}  

\title{Domain Adaptive Object Detection via Asymmetric Tri-way Faster-RCNN} 

\titlerunning{Domain Adaptive Object Detection via Asymmetric Tri-way Faster-RCNN}
%
\author{Zhenwei He\orcidID{0000-0002-6122-9277} \and
Lei Zhang\textsuperscript{(\Letter)}\orcidID{0000-0002-5305-8543}}
\authorrunning{Zhenwei He and Lei Zhang}
%
\institute{Learning Intelligence \& Vision Essential (LiVE) Group \\ School of Microelectronics and Communication Engineering, Chongqing University
\email{\{hzw,leizhang\}@cqu.edu.cn}\\
\url{http://www.leizhang.tk/}}
\maketitle

\begin{abstract}
Conventional object detection models inevitably encounter a performance drop as the domain disparity exists. Unsupervised domain adaptive object detection is proposed recently to reduce the disparity between domains, where the source domain is label-rich while the target domain is label-agnostic. The existing models follow a parameter shared siamese structure for adversarial domain alignment, which, however, easily leads to the collapse and out-of-control risk of the source domain and brings negative impact to feature adaption. The main reason is that the labeling unfairness (asymmetry) between source and target makes the parameter sharing mechanism unable to adapt. Therefore, in order to avoid the source domain collapse risk caused by parameter sharing, we propose an asymmetric tri-way Faster-RCNN (ATF) for domain adaptive object detection. Our ATF model has two distinct merits: 1) A ancillary net supervised by source label is deployed to learn ancillary target features and simultaneously preserve the discrimination of source domain, which enhances the structural discrimination (object classification vs. bounding box regression) of domain alignment. 2) The asymmetric structure consisting of a chief net and an independent ancillary net essentially overcomes the parameter sharing aroused source risk collapse. The adaption safety of the proposed ATF detector is guaranteed. Extensive experiments on a number of datasets, including Cityscapes, Foggy-cityscapes, KITTI, Sim10k, Pascal VOC, Clipart and Watercolor, demonstrate the SOTA performance of our method.

\keywords{Object detection; Transfer learning; Deep learning}
\end{abstract}

\section{Introduction}

Object detection is one of the significant tasks in computer vision, which has extensive applications in video surveillance, self-driving, face analysis, medical imaging, etc. Motivated by the development of CNNs, researchers have made the object detection models fast, reliable, and precise~\cite{girshick2015fast,he2018mask,ren2017faster,liu2016ssd,fu2017dssd,lin2017focal}. However, in real-world scenarios, object detection models face enormous challenges due to the diversity of application environments such as different weathers, backgrounds, scenes, illuminations, and object appearances. The unavoidable environment change causes the domain shift circumstance for object detection models. Nevertheless, conventional object detection models are domain constrained, which can not take into account the domain shift happened in open conditions and noticeable performance degradation is resulted due to the environmental changes. One way to avoid the influence of domain shift is to train the detector by domain/scene specific data, but labeling the training data of each domain is time-consuming and impractical. For reducing the labeling cast and getting a domain adaptive detector, in this paper, the unsupervised domain adaptive object detection task is addressed by transferring the label-rich source domain to the label-agnostic target domain. Since the training label of the target domain is unnecessary, there is no extra annotation cast for the target domain. More importantly, detectors are more generalizable to the environmental change benefitting from the co-training between domains.

\begin{figure}[t]
\centering
\subfigure[Enhancement of Classifier]
{\includegraphics[width=0.52\linewidth]{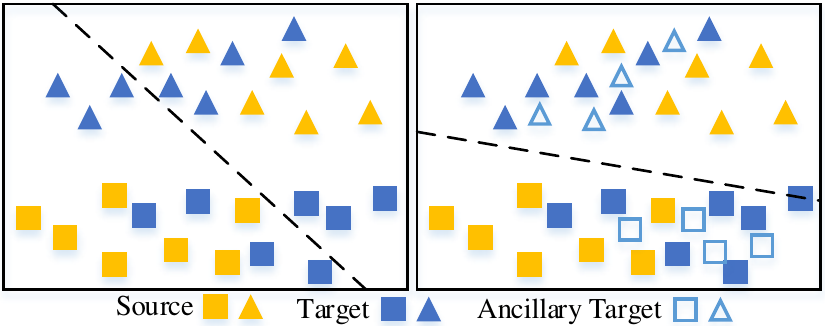}}
\subfigure[Accuracy on Target Domain]
{\includegraphics[width=0.45\linewidth]{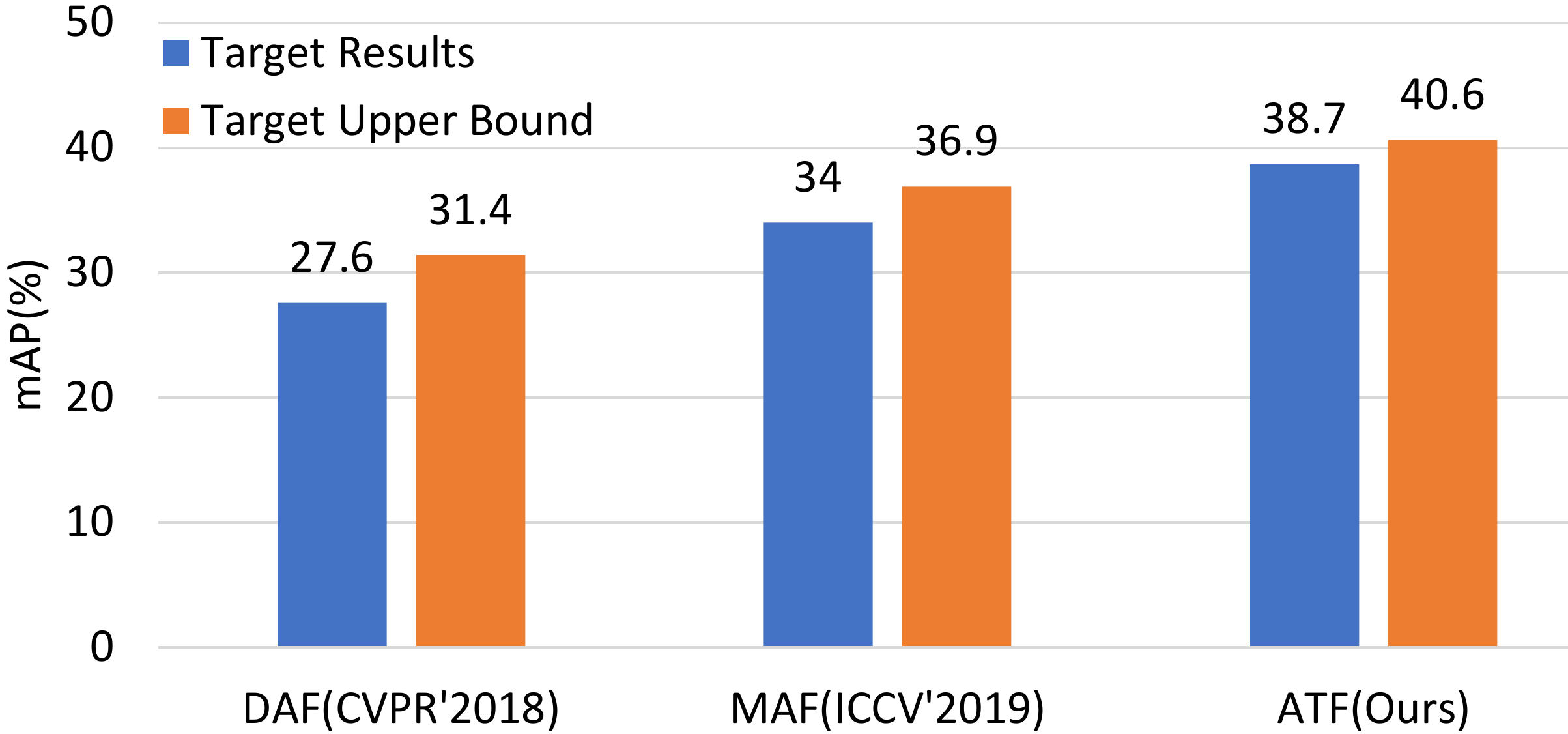}}
\caption{The motivation of our approach. (a) shows the learning effect of the classification decision boundary for both source and target data without/with the ancillary target features. (b) presents the target domain detection performance (mAP) based on different transfer models and the upper bound of the target performance based on their deep features and groundtruth target labels.}
\label{fig:introduction}
\end{figure}

Very recently, unsupervised domain adaptive object detection is proposed to mitigate the domain shift problem by transferring the knowledge from the sematic related source domain to target domain~\cite{chen2018domain,he2019multi,kim2019diversify,khodabandeh2019robust}. Most of cross-domain object detection models learn the domain invariant features with transfer learning ideas. Inspired by the pioneer of this field~\cite{chen2018domain,he2019multi}, the detector can be domain-invariant if only the source and target domains are sufficiently confused. No suspicion that a domain-invariant feature representation can enable the object detector to be domain adaptive. However, the domain-invariant detector does not guarantee good object classification and bounding box (bbox) regression. This is mainly due to the lack of domain specific data. Relying solely on labeled source data and unlabeled target data to entirely eliminate the domain disparity is actually not an easy task. This is motivated in Fig.~\ref{fig:introduction}(a) (left), where we use a toy classifier as an example. We can see that as the features from different domains are not fully aligned, the decision boundary learned with labeled source data and unlabeled target data at hand can not correctly classify the samples from the target domain. Thus, we have an instinct thought to implicitly learn ancillary target data for domain-invariant class discriminative and bbox regressive features. Specifically, to better characterize the domain-invariant detector, as is shown in Fig.~\ref{fig:introduction}(a) (right), we propose to learn the ancillary target features with a specialized module, which we call ``ancillary net''. We see that the target process features contribute to the new decision boundary of the classifier such that the target features are not only domain invariant but class separable.

Most of the CNN based domain adaption algorithms aim to learn a transferable feature representation~\cite{saito2018open,xu2019larger,chen2019crdoco,ganin2016domain}. They utilize the feature learning ability of CNN to extract domain-invariant representation for the source and target domains. A popular strategy to learn transferable features with CNN is adversarial training just like the generative adversarial net (GAN)~\cite{goodfellow2014generative}. The adversarial training strategy is a two-player gaming, in which a discriminator is trained to distinguish different domains, while a generator is trained to extract domain-invariant features to fool the discriminator. However, adversarial learning also has some risks. As indicated by~\cite{liu2019transferable}, forcing the features to be domain-invariant may inevitably distort the original distribution of domain data, and the structural discrimination (intra-class compactness vs. inter-class separability) between two domains may be destroyed. This is mainly because the target data is completely unlabeled.

\begin{figure}[t]
\centering
\includegraphics[width=1.0\linewidth]{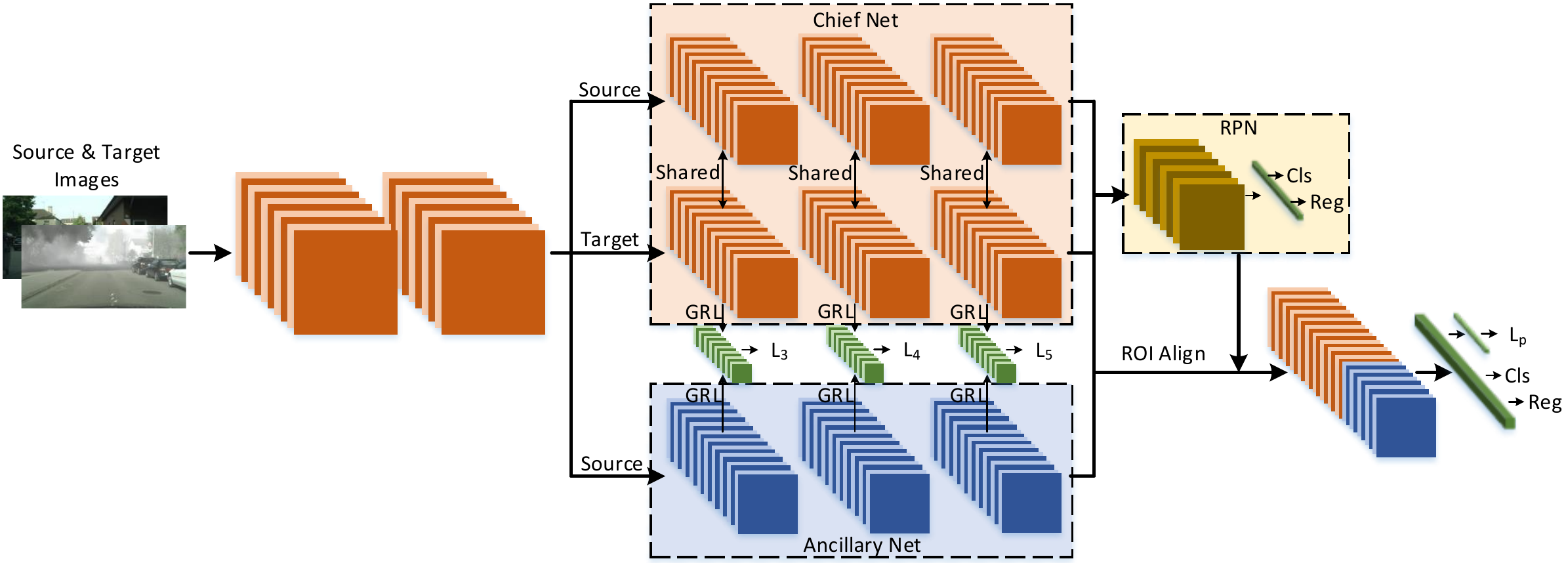}
\caption{The net structure of our ATF. Our ATF has three streams upon the backbone. The first two streams with shared parameters are the Chief net (orange color). Another stream is the ancillary net (blue color) which is independent of the chief net. All streams are fed into the same RPN. A ROI-Align layer is deployed to get the pooled features of all streams. The pooled features are used for final detection. The chief net is trained by ancillary data guided domain adversarial loss and source label guided detection loss. The ancillary net is trained with source label guided detection loss.}
\label{fig:strucuture}
\end{figure}

Similarly, distribution distortion also occurs in cross-domain object detection. Since the target data is unlabeled and the model is trained only with source labels, the learned source features can be discriminative and reliable, while the discrimination of the target features is vulnerable and untrustworthy. However, most existing models such as DAF~\cite{chen2018domain} and MAF~\cite{he2019multi} default that the source and target domains share the same network with parameter sharing. A forthcoming problem of parameter sharing network is that aligning the reliable source features toward the unreliable target features may enhance the risk of source domain collapse and eventually deteriorate the structural discrimination of the model. It will inevitably bring a negative impact to object classification and bbox regression of the detector. According to the domain adaption theory in~\cite{Bendavid2006Analysis}, the expected target risk $\epsilon_T(h)$ is upper bounded by the empirical source risk $\epsilon_S(h)$, domain discrepancy $d_\mathcal{A}$ and shared error $\lambda=\epsilon_T(h^*)+\epsilon_S(h^*)$ of the ideal hypothesis $h^*$ for both domains. Therefore, effectively controlling the source risk and avoiding the collapse of the source domain is particularly important for the success of a domain adaptive detector. In this paper, we propose an \textbf{A}symmetric \textbf{T}ri-way structure to enhance the transferability of \textbf{F}aster-RCNN, which is called ATF and consists of a chief net and an ancillary net, as is shown in Fig.~\ref{fig:strucuture}. The asymmetry originates from that the ancillary net is independent of the parameter shared chief net. Because the independent ancillary net is only trained by the labeled source data, the asymmetry can largely avoid source collapse and feature distortion during transfer.

Our model inclines to preserve the discrimination of source features and simultaneously guide the structural transfer of target features. One evidence is shown in Fig.~\ref{fig:introduction} (b), in which we implement the domain adaptive detectors, such as DAF~\cite{chen2018domain}, MAF~\cite{he2019multi} and ATF from Cityscapes~\cite{cordts2016cityscapes} dataset to the Foggy Cityscapes~\cite{sakaridis2018semantic}, respectively. Additionally, in order to observe the upper target performance, we also use the features from the networks and train the detector with the ground truth target label. We can see that our ATF achieves both higher performance (11.1\% and 4.7\% resp.) than DAF (CVPR'18) and MAF (ICCV'19).
In summary, this paper has two distinct merits. 1) We propose a source only guided ancillary net in order to learn ancillary target data for reducing the domain bias and model bias. 2) We propose an asymmetric tri-way structure, which overcomes the adversarial model collapse of the parameter shared network, i.e. the out-of-control risk in the source domain.

\section{Related Work}

\textbf{Object detection}. Object detection is an essential task of computer vision, which has been studied for many years. Boosted by the development of deep convolutional neural networks, object detection has recently achieved significant advances~\cite{lin2017feature,cai2018cascade,redmon2018yolov3,vu2019cascade}. Object detection models can be roughly categorized into two types: one-stage and two-stage detection. One-stage object detection is good at computational efficiency, which attains real-time object detection. SSD~\cite{liu2016ssd} is the first one-stage object detector. Although SSD has a similar way to detect objects as RPN~\cite{ren2017faster}, it uses multiple layers for various scales. YOLO~\cite{redmon2016you} outputs sparse detection results with high computation speed. Recently, RetinaDet~\cite{lin2017focal} addresses the unbalance of foreground and background with the proposed focal loss. Two-stage detectors generate region proposals for the detection, for example, Faster-RCNN~\cite{ren2017faster} introduced the RPN for the proposal generation. FPN~\cite{lin2017feature} utilized multi-layers for the detection of different scales. NAS-FPN~\cite{ghiasi2019fpn} learn the net structure of FPN to achieve better performance. In this paper, we select the Faster-RCNN as the base network for domain adaptive detection.

\textbf{Domain Adaption}. Domain adaption is proposed for bridging the gap between different domains. The domain adaption problem has been investigated for several different computer vision tasks, such as image classification and object segmentation~\cite{saito2018open,saito2018maximum,wang2019unsupervised,chen2019crdoco}. Inspired by the accomplishment of deep learning, early domain adaptive models minimize the disparity estimate between different domains such as maximum mean discrepancy (MMD)~\cite{long2016unsupervised,sun2016deep}. Recently, several domain adaptive models were proposed based on adversarial learning. The two-player game between the feature extractor and domain discriminator promotes the confusion between different domains. Ganin~\emph{et al}.~\cite{ganin2014unsupervised} proposed the gradient reverse layer (GRL), which reverses the gradient backpropagation for adversarial learning. For feature regularization, Cicek~\emph{et al}. introduced a regularization method base on the VAT. Besides the adversarial of extractor and discriminator, Saito~\emph{et al}.~\cite{saito2018maximum} employed two distinct classifiers to reduce the domain disparity. In our ATF, an unsupervised domain adaption mechanism is exploited.

\textbf{Domain Adaptive Object Detection}. Domain adaptive or cross-domain object detection has been raised very recently for the unconstrained scene. This task was firstly proposed by Chen~\emph{et al}~\cite{chen2018domain}, which addresses the domain shift on both image-level and instance-level. This work indicated that if the features are domain-invariant, the detector can achieve better performance on the target domain. After that, several models are also proposed for domain adaptive detection. In~\cite{saito2019strong}, low-level features and high-level features are treated with strong and weak alignment, respectively. Kim~\emph{et al}.~\cite{kim2019self} introduces a new loss function and exploits the self-training strategy to improve the detection of the target domain. He and Zhang proposed a MAF~\cite{he2019multi} model, in which a hierarchical structure is designed to reduce the domain disparity at different scales. In addition to the domain adaption guided detectors, the mean teacher is introduced to get the pseudo labels for the target domain~\cite{cai2019exploring}. Besides that, Khodabandeh~\emph{et al}.~\cite{khodabandeh2019robust} train an extra classifier and use KL-divergence to learn a model for correction.

\section{The Proposed ATF Approach}

In this section, we introduce the details of our Asymmetric Tri-way Faster-RCNN (ATF) model. For convenience, the fully labeled source domain is marked as $D_{s}=\left\{(x_{i}^{s}, b_{i}^{s}, y_{i}^{s}) \right\}_{i}^{n_{s}}$, where $x_{i}^{s}$ stands the image, $b_{i}^{s}$ is the coordinate of bounding boxes, $y_{i}^{s}$ is the category label and $n_{s}$ is the number of samples. The unlabeled target domain is marked as $D_{t}=\left\{(x_{i}^{t}) \right\}_{i}^{n_{t}}$, where $n_{t}$ denotes the number of samples. Our task is to transfer the semantic knowledge from $D_{s}$ to $D_{t}$ and achieve successful detection in the target domain.

\subsection{Network Architecture of ATF}

Our proposed ATF model is based on the Faster-RCNN~\cite{ren2017faster} detection framework. In order to overcome the out-of-control risk of source domain in the conventional symmetric Siamese network structure with shared parameters, we introduce an asymmetric tri-way network as the backbone. Specifically, the images from the source or target domain are fed into the first two convolution blocks. On top of that, we divide the structure into three streams. As shown in Fig.~\ref{fig:strucuture}, the first two streams with the shared parameters in orange color are the Chief net. Features from the source and target data are fed into the two streams, respectively. The third stream with blue color is the proposed Ancillary net, which is parametrically independent of the chief net. That is, the ancillary net has different parameters from the chief net. The source only data is fed into the ancillary net during the training phase. Three streams of the network share the same region proposal network (RPN) as Faster-RCNN does. We pool the features of all streams based on the proposals with the ROI-Align layer. Finally, we get the detection results on top of the network with the pooled features.

An overview of our network structure is illustrated in Fig.~\ref{fig:strucuture}. For training the ATF model, we design the adversarial domain confusion strategy, which is established between the chief net and ancillary net. The training loss of our ATF consists of two kinds of losses: domain adversarial confusion (\textbf{Dac}) loss in the chief net for bounding the domain discrepancy $d_{\mathcal{A}}$ and the source labeled guided detection loss (\textbf{Det}) in the ancillary net for bounding the empirical source risk $\epsilon_S(h)$. So, the proposed model is easily trained end-to-end.

\subsection{Principle of the Chief Net}

Domain discrepancy is the primary factor that leads to performance degradation in cross-domain object detection. In order to reduce the domain discrepancy $d_{\mathcal{A}}$, we introduce the domain adversarial confusion (Dac) mechanism which bridges the gap between the chief net (target knowledge) and the ancillary net (source knowledge). The features from the ancillary net should have a similar distribution to the target stream features from the chief net. Considering that object detection refers to two stages, i.e., image-level feature learning (global) and proposal-level feature learning (local), we propose two alignment modules based on the Dac mechanism, i.e. global domain alignment with Dac and local domain alignment with Dac.

\textbf{Global domain alignment with Dac}. Obviously, the Dac guided global domain alignment focuses on the low-level convolutional blocks between the chief net (target stream) and the ancillary net in ATF. Let $x_{i}^{s}$ and $x_{i}^{t}$ be two images from the source and target domains, respectively. The feature maps of the $k^{th} (k=3, 4, 5)$ block of the chief net and ancillary net as shown in Fig.~\ref{fig:strucuture} are defined as $F_{c}(x_{i}^{t},\theta_{c}^{k})$ and $F_{a}(x_{i}^{s},\theta_{a}^{k})$, respectively. $d$ is the binary domain label, and $d=1$ for source domain and 0 for target domain. The Dac based global domain alignment loss for $k^{th}$ block ($\mathcal{L}^k_{G-Dac}$) can be written as:
\begin{equation}
\mathcal{L}_{G-Dac}^k=-\sum _{u,v}((1-d)\log(D_{k}(F_{c}(x_{i}^{t},\theta_{c}^{k})^{(u,v)},\theta_{d}^{k})+d\log(D_{k}(F_{a}(x_{i}^{s},\theta_{a}^{k})^{(u,v)},\theta_{d}^{k}))
\label{eq1}
\end{equation}
where the $(u,v)$ stands for the pixel coordinate of the feature map. $D_{k}$ is the discriminator of $k^{th}$ block. $\theta_{c}^{k}$, $\theta_{a}^{k}$, and $\theta_{d}^{k}$ are the parameters of chief net, ancillary net and discriminator in the $k^{th}$ block, respectively. In principle, the discriminator $D$ tries to minimize $\mathcal{L}_{G-Dac}^k(D)$ to distinguish the features from different domains. With the gradient reversal layer (GRL)~\cite{ganin2014unsupervised}, the chief net $F_c$ and ancillary net $F_a$ try to maximize $\mathcal{L}_{G-Dac}^k(F_c, F_a)$. Then, the global features between the chief net (target stream) and the ancillary net are confused (aligned).

\textbf{Local domain alignment with Dac}. The domain alignment on the global image level is still not enough for the local object based detector. Therefore, we propose to further align the local object-level features across domains. As shown in Fig.~\ref{fig:strucuture}, the features pooled by the ROI-Align layer stand for the local part of an image, including foreground and background. Similar to the global domain alignment module, we confuse (align) the local object-level features pooled from the target stream of the chief net and the ancillary net. Suppose the pooled features from the ancillary net to be $f_{a}$ and features from the chief net to be $f_{c}$, the Dac based local domain alignment loss ($\mathcal{L}_{L-Dac}$) on the local object-level features is formulated as:
\begin{equation}
\mathcal{L}_{L-Dac}=-\frac{1}{N}\sum_{n}((1-d)\log(D_{l}(F_{l}(f_{c}^{n},\theta_{f}),\theta_{d})+d\log(D_{l}(F_{l}(f_{a}^{n},\theta_{f}),\theta_{d}))
\label{eq2}
\end{equation}
where the $D_{l}$ is the local domain discriminator, $F_{l}$ is the local backbone network, $\theta_{l}$ and $\theta_{d}$ are the parameters of the backbone and the discriminator, respectively. We implement the adversarial learning with GRL~\cite{ganin2014unsupervised}. The discriminator tries to minimize $\mathcal{L}_{L-Dac}(D_{l})$ while the local backbone network tries to maximize $\mathcal{L}_{L-Dac}(F_{l})$ for local domain confusion.

\subsection{Principle of the Ancillary net}

As discussed above, the chief net aims to bound the domain discrepancy $d_{A}$. In this section, we propose to bound the empirical source risk $\epsilon_S$ by using the ancillary net. The reason why we construct a specialized ancillary net for bounding the source risk rather than using the source stream of the chief net has been elaborated. The main reason is that the chief net is parameter shared, so the empirical risk of source stream in the chief net is easily out-of-control due to the unlabeled problem of the target domain. Because the source domain is sufficiently labeled with object categories and bounding boxes in each image, the source risk $\epsilon_S$ of the ancillary net is easy to be bounded by the classification loss and regression loss of detector. In our implementation, the detection loss for the chief net is reused for the supervision of the ancillary net.

From the principles of the chief net and the ancillary net, we know that the ancillary net is trained to generate features that have a similar distribution to the target stream in the chief net as shown in Eqs.(\ref{eq1}) and (\ref{eq2}). That is, the ancillary net adjusts the features learned by the target stream of the chief net to adapt to the source data trained detector. Meanwhile, the ancillary net is restricted by the classifier and regressor of the source detector, such that the structural discrimination is preserved in the source domain. Therefore, with the domain alignment and source risk minimization, the expected task risk can be effectively bounded for domain adaptive object detection.

\subsection{Training Loss of Our ATF}

The proposed asymmetric tri-way Faster-RCNN (ATF) contains the two loss functions, the detection based source risk loss and the domain alignment loss. The detection loss function for both chief and ancillary nets is shown as:
\begin{equation}
\mathcal{L}_{Det}=\mathcal{L}_{cls}(x_{i}^{s},b_{i}^{s},y_{i}^{s})+\mathcal{L}_{reg}(x_{i}^{s},b_{i}^{s},y_{i}^{s})
\label{eq3}
\end{equation}
where $\mathcal{L}_{cls}$ is the softmax based cross-entropy loss and $\mathcal{L}_{reg}$ is the smooth-$L_1$ loss, which are standard detection losses for bounding the empirical source risk. In summary, by revisiting the Eqs.(\ref{eq1}), (\ref{eq2}) and (\ref{eq3})), the total loss function for training our model can be written as:
\begin{equation}
\mathcal{L}_{ATF}=\mathcal{L}_{Det}+\alpha(\mathcal{L}_{L-Dac}+\sum_{k=3}^{5}\mathcal{L}_{G-Dac}^{k})
\label{eq4}
\end{equation}
Where the $\alpha$ is a hyper-parameter to adjust the weight of domain alignment loss. The model is easily trained end-to-end with Stochastic Gradient Descent (SGD). Overview of our ATF can be observed in Fig.~\ref{fig:strucuture}.

\section{Experiments}

In this section, we evaluate our approach on several different datasets, including Cityscapes~\cite{cordts2016cityscapes}, Foggy Cityscapes~\cite{sakaridis2018semantic}, KITTI~\cite{geiger2012we}, SIM10k~\cite{johnson2016driving}, Pascal VOC~\cite{everingham2010pascal}, Clipart~\cite{inoue2018cross} and Watercolor~\cite{inoue2018cross}. We compare our results with state-of-the-art methods to show the effectiveness of our model.

\subsection{Implementation Details}

The base network of our ATF model is VGG-16~\cite{simonyan2014very} or ResNet-101~\cite{he2016deep} in the experiments. We follow the same experimental setting as~\cite{chen2018domain}, where the source domain is fully labeled, and the target domain is completely unlabeled. The trade-off parameter $\alpha$ in Eq.(\ref{eq3}) is set as 0.7 in our implementation. We use the ImageNet pre-trained model for the initialization of our model. For each iteration, one labeled source sample and one unlabeled target sample are fed into ATF for training. In the test phase, the well-trained chief net is used to get the detection results. For all datasets, we report the average precisions (AP, \%) and mean average precisions (mAP, \%) with a threshold of 0.5.

\subsection{Datasets}

\textbf{Cityscapes:} Cityscapes~\cite{cordts2016cityscapes} captures high-quality video for the outdoor scenes in different cities for automotive vision. The dataset includes 5000 manually selected images from 27 cities, which are collected with a similar weather condition. These images are annotated with dense pixel-level image annotation. Although the Cityscapes dataset is labeled for the semantic segmentation task, we generate the bounding box based on the pixel-level label in the experiment as~\cite{chen2018domain} did for the cross-domain detection task.

\textbf{Foggy Cityscapes:} Foggy Cityscapes~\cite{sakaridis2018semantic} dataset simulates the foggy weather based on the Cityscapes. The pixel-level labels of Cityscapes can be inherited by the Foggy Cityscapes such that we can generate the bounding box for the dataset. In the experiments, the validation set of the Foggy Cityscapes is used as the testing set in our experiment.

\textbf{KITTI:} KITTI~\cite{geiger2012we} dataset is collected by the autonomous driving platform, which includes two color and two grayscale PointGrey Flea2 video cameras. Images of the dataset are manually selected in several different scenes in a mid-sized city. The dataset includes 14999 images and 80256 bounding boxes for the detection task. Only the training set of KITTI is used for our experiment.

\textbf{SIM10K:} Images of SIM10K~\cite{johnson2016driving} are generated by the engine of Grand Theft Auto V (GTA V). The dataset simulates different scenes, such as different time or weather. SIM10K contains 10000 images with 58701 bounding boxes of car. All images of the dataset are used for training.

\textbf{Pascal VOC:} Pascal VOC~\cite{everingham2010pascal} is a famous object detection dataset. This dataset contains 20 categories with bounding boxes. The image scale of the dataset is diverse. In our experiment, the training and validation split of VOC07 and 12 are used as the training set, which results in about 15k images.

\textbf{Clipart and Watercolor:} The Clipart and Watercolor~\cite{inoue2018cross} are constructed by the Amazon Mechanical Turk, which is introduced for the domain adaption detection task. Similar to the Pascal VOC, the Clipart contains 1000 images and 20 categories. Watercolor has 2000 images of 6 categories. Half of the datasets are introduced for training while the remaining is used for the test.

\subsection{Cross-domain Detection in Different Visibility and Cameras}

\textbf{Domain adaption across different visibility}. Visibility change caused by weather can shift the data distribution of the collected images. In some application scenarios, such as autonomous driving, the detection model has to adapt to different weather conditions. In this section, we evaluate our ATF with the cityscapes~\cite{cordts2016cityscapes} and the foggy cityscapes~\cite{sakaridis2018semantic} datasets. We treat the Cityscapes as the source domain and the Foggy Cityscapes as the target domain. Our model uses VGG16 as the base net in the experiment. We introduce the source only trained Faster-RCNN (without adaptation), DAF~\cite{chen2018domain}, MAF~\cite{he2019multi}, Strong-Weak~\cite{saito2019strong}, Diversify\&match(D\&match)~\cite{kim2019diversify}, Noisy Labeling(NL)~\cite{khodabandeh2019robust}, and SCL~\cite{shen2019scl} for the comparison. Our ATF is trained for 18 epochs in the experiment, where the learning rate is set as 0.001 and changes to 0.0001 in the 12$^{th}$ epoch.

\begin{table}[t]
\begin{center}
\caption{The cross-domain detection results from Cityscapes to Foggy Cityscapes.}
\label{table:f_city}
\begin{tabular}{p{3cm}|p{0.9cm}<{\centering}p{0.9cm}<{\centering}p{0.9cm}<{\centering}p{0.9cm}<{\centering}p{0.9cm}<{\centering}p{0.9cm}<{\centering}p{0.9cm}<{\centering}p{0.9cm}<{\centering}||p{0.9cm}<{\centering}}
\hline
Methods & person & rider & car & truck & bus & train & mcycle & bcycle & mAP \\
\hline
\hline
Faster-RCNN & 24.1 & 33.1 & 34.3 & 4.1 & 22.3 & 3.0 & 15.3 & 26.5 & 20.3 \\
DAF(CVPR'18)~\cite{chen2018domain} & 25.0 & 31.0 & 40.5 & 22.1 & 35.3 & 20.2 & 20.0 & 27.1 & 27.6 \\
MAF(ICCV'19)~\cite{he2019multi} & 28.2 & 39.5 & 43.9 & 23.8 & 39.9 & 33.3 & 29.2 & 33.9 & 34.0 \\
Strong-Weak~\cite{saito2019strong} & 29.9 & 42.3 & 43.5 & 24.5 & 36.2 & 32.6 & 30.0 & 35.3 & 34.3 \\
D\&match~\cite{kim2019diversify} & 30.8 & 40.5 & 44.3 & 27.2 & 38.4 & 34.5 & 28.4 & 32.2 & 34.6 \\
NL /w res101~\cite{khodabandeh2019robust} & \textbf{35.1} & 42.2 & 49.2 & 30.1 & 45.3 & 27.0 & 26.9 & 36.0 & 36.5 \\
SCL~\cite{shen2019scl} & 31.6 & 44.0 & 44.8 & \textbf{30.4} & 41.8 & \textbf{40.7} & \textbf{33.6} & 36.2 & 37.9 \\
\hline
\hline
ATF(1-block) & 33.3 & 43.6 & 44.6 & 24.3 & 39.6 & 10.5 & 27.2 & 35.6 & 32.3 \\
ATF(2-blocks) & 34.0 & 46.0 & 49.1 & 26.4 & \textbf{46.5} & 14.7 & 30.7 & 37.5 & 35.6 \\
ATF(ours) & 34.6 & \textbf{47.0} & \textbf{50.0} & 23.7 & 43.3 & 38.7 & 33.4 & \textbf{38.8} & \textbf{38.7} \\
ATF* & 34.6 & 46.5 & 49.2 & 23.5 & 43.1 & 29.2 & 33.2 & 39.0 & 37.3 \\
\hline
\end{tabular}
\end{center}
\end{table}

The results are presented in Table~\ref{table:f_city}. We can see that our ATF achieves 38.7\% mAP, which outperforms all the compared models. Due to the lack of domain specific data, the models which only concentrate on the feature alignment can not work well, such as MAF~\cite{he2019multi}, Strong-Weak~\cite{saito2019strong} and SCL~\cite{shen2019scl}. With the ancillary target features from the ancillary net to reduce the domain shift and bias, our model gets preferable performance. Additionally, our ATF model also outperforms the pseudo label based model~\cite{khodabandeh2019robust}, in which it has to generate and update the pseudo labels with features extracted by a source only trained extractor, where the target features are untrustworthy. The unreliable feature based pseudo labels can not lead to a precise target model. In order to prove the effectiveness of ancillary net, we conduct ablation studies that reduce the convolutional blocks of ancillary net. The results of 1-block (the $3^{rd}$ and $4^{th}$ blocks are removed) and 2-blocks (the $3^{rd}$ blocks is removed) from the ancillary net are shown in Table~\ref{table:f_city}. As we reduce the convolutional block of the ancillary net, the performance drops. Otherwise, we freeze the parameter of the backbone and fine-tune the regressors and classifiers of the ATF detector with source labels. The performance of the fine-tuned model denoted as ATF* is also presented in Table~\ref{table:f_city}. We find that the performance of ATF* is inferior to the ATF because the effect of ancillary target features computed by the ancillary net is removed. The merit of the proposed ancillary net is validated.
\begin{table}[h]
\begin{center}
\caption{The results of domain adaptive object detection on Cityscapes and KITTI.}
\label{table:f_kc}
\begin{tabular}{|p{1.1cm}|p{2cm}|p{1.5cm}|p{1.5cm}|p{1.5cm}|p{1.5cm}|p{1.5cm}|}
\hline
Tasks& Faster-RCNN & DAF~\cite{chen2018domain} & MAF~\cite{he2019multi} & S-W~\cite{saito2019strong} & SCL~\cite{shen2019scl} & ATF(ours) \\
\hline
\hline
K to C & 30.2 & 38.5 & 41.0 & 37.9 & 41.9 & \textbf{42.1} \\
\hline
C to K & 53.5 & 64.1 & 72.1 & 71.0 & 72.7 & \textbf{73.5} \\
\hline
\end{tabular}
\end{center}
\end{table}

\textbf{Domain adaption across different cameras}. The camera change is another important factor leading to the domain shift in real-world application scenarios. In this experiment, we employ the Cityscapes (C)~\cite{sakaridis2018semantic} and KITTI (K)~\cite{geiger2012we} as the source and target domains, respectively. The source only trained Faster-RCNN (without adaption), DAF~\cite{chen2018domain}, MAF~\cite{he2019multi}, and Strong-Weak~\cite{saito2019strong} are implemented for comparisons. The AP of car on the target domain is computed for the test. The experimental results are presented in Table~\ref{table:f_kc}.

In Table~\ref{table:f_kc}, K to C represents that the KITTI is used as the source domain, while the Cityscapes is used as the target domain and vice versa. We can observe that our ATF achieves the best performance on both K to C and C to K tasks among all the compared models, which testify the effectiveness of our model in alleviating the domain shift problem caused by the change of cameras.

\subsection{Cross-domain Detection on Large Domain Shift}

\begin{table}[t]
\begin{center}
\caption{The cross-domain detection results from Pascal VOC to Clipart.}
\label{table:clipart}
\begin{tabular}{l|p{0.8cm}<{\centering}p{0.8cm}<{\centering}p{0.8cm}<{\centering}p{0.8cm}<{\centering}p{0.8cm}<{\centering}p{0.8cm}<{\centering}p{0.8cm}<{\centering}p{0.8cm}<{\centering}p{0.8cm}<{\centering}p{0.8cm}<{\centering}|c}
\hline
Methods & aero & bike & bird & boat & bottle & bus & car & cat & chair & cow & ~ \\
\hline
\hline
Faster-RCNN & 35.6 & 52.5 & 24.3 & 23.0 & 20.0 & 43.9 & 32.8 & 10.7 & 30.6 & 11.7 & ~ \\
DAF~\cite{chen2018domain} & 15.0 & 34.6 & 12.4 & 11.9 & 19.8 & 21.1 & 23.2 & 3.1 & 22.1 & 26.3 & ~ \\
BDC-Faster & 20.2 & 46.4 & 20.4 & 19.3 & 18.7 & 41.3 & 26.5 & 6.4 & 33.2 & 11.7 & ~ \\
\hline
WST-BSR~\cite{kim2019self} & 28.0 & 64.5 & 23.9 & 19.0 & 21.9 & \textbf{64.3} & \textbf{43.5} & 16.4 & \textbf{42.2} & 25.9 & ~ \\
Strong-Weak~\cite{saito2019strong} & 26.2 & 48.5 & 32.6 & 33.7 & 38.5 & 54.3 & 37.1 & 18.6 & 34.8 & 58.3 & ~ \\
MAF~\cite{he2019multi} & 38.1 & 61.1 & 25.8 & \textbf{43.9} & 40.3 & 41.6 & 40.3 & 9.2 & 37.1 & 48.4 & ~ \\
SCL~\cite{shen2019scl} & \textbf{44.7} & 50.0 & \textbf{33.6} & 27.4 & \textbf{42.2} & 55.6 & 38.3 & \textbf{19.2} & 37.9 & 69.0 & ~ \\
\hline
ATF(ours) & 41.9 & \textbf{67.0} & 27.4 & 36.4 & 41.0 & 48.5 & 42.0 & 13.1 & 39.2 & \textbf{75.1} & ~ \\
\hline
\hline
Methods & table & dog & horse & mbike & prsn & plant & sheep & sofa & train & tv & mAP \\\hline
\hline
Faster-RCNN & 13.8 & 6.0 & 36.8 & 45.9 & 48.7 & 41.9 & 16.5 & 7.3 & 22.9 & 32.0 & 27.8 \\
DAF~\cite{chen2018domain} & 10.6 & 10.0 & 19.6 & 39.4 & 34.6 & 29.3 & 1.0 & 17.1 & 19.7 & 24.8 & 19.8 \\
BDC-Faster & 26.0 & 1.7 & 36.6 & 41.5 & 37.7 & 44.5 & 10.6 & 20.4 & 33.3 & 15.5 & 25.6 \\
\hline
WST-BSR~\cite{kim2019self} & 30.5 & 7.9 & 25.5 & \textbf{67.6} & 54.5 & 36.4 & 10.3 & \textbf{31.2} & \textbf{57.4} & 43.5 & 35.7 \\
Strong-Weak~\cite{saito2019strong} & 17.0 & 12.5 & 33.8 & 65.5 & \textbf{61.6} & 52.0 & 9.3 & 24.9 & 54.1 & \textbf{49.1} & 38.1 \\
MAF~\cite{he2019multi} & 24.2 & 13.4 & 36.4 & 52.7 & 57.0 & \textbf{52.5} & 18.2 & 24.3 & 32.9 & 39.3 & 36.8 \\
SCL~\cite{shen2019scl} & 30.1 & \textbf{26.3} & 34.4 & 67.3 & 61.0 & 47.9 & 21.4 & 26.3 & 50.1 & 47.3 & 41.5 \\
\hline
ATF(ours) & \textbf{33.4} & 7.9 & \textbf{41.2} & 56.2 & 61.4 & 50.6 & \textbf{42.0} & 25.0 & 53.1 & 39.1 & \textbf{42.1} \\
\hline
\end{tabular}
\end{center}
\end{table}

In this section, we concentrate on the domains with large domain disparity, especially, from the real image to the comical or artistic images. We employ the Pascal VOC~\cite{everingham2010pascal} dataset as the source domain, which contains the real image. The Clipart or Watercolor~\cite{inoue2018cross} is exploited as the target domain. The backbone for the experiments is the ImageNet pretrained ResNet-101. We train our model for 8 epochs with the learning rate of 0.001 and change the learning rate to 0.0001 in the 6$^{th}$ epoch to ensure convergence.

\textbf{Transfer from Pascal VOC to Clipart}. The Clipart~\cite{inoue2018cross} contains the comical images which has the same 20 categories as Pascal VOC~\cite{everingham2010pascal}. In this experiment, we introduce the source only Faster RCNN, DAF~\cite{chen2018domain}, WST-BSR~\cite{kim2019self}, MAF~\cite{he2019multi}, Strong-Weak~\cite{saito2019strong} and SCL~\cite{shen2019scl} for the comparison. The results are shown in Table~\ref{table:clipart}. Our ATF achieves 42.1\% mAP and outperforms all models.

\begin{table}[h]
\begin{center}
\caption{The cross-domain detection results from Pascal VOC to Watercolor.}
\label{table:water}
\begin{tabular}{|p{2.5cm}|p{1cm}<{\centering}p{1cm}<{\centering}p{1cm}<{\centering}p{1cm}<{\centering}p{1cm}<{\centering}p{1cm}<{\centering}||p{1cm}<{\centering}|}
\hline
Methods & bike & bird & car & cat & dog & person & mAP \\
\hline
\hline
Faster-RCNN & 68.8 & 46.8 & 37.2 & 32.7 & 21.3 & 60.7 & 44.6 \\
DAF~\cite{chen2018domain} & 75.2 & 40.6 & 48.0 & 31.5 & 20.6 & 60.0 & 46.0 \\
BDC-Faster & 68.6 & 48.3 & 47.2 & 26.5 & 21.7 & 60.5 & 45.5 \\
WST-BSR~\cite{kim2019self} & 75.6 & 45.8 & \textbf{49.3} & 34.1 & 30.3 & 64.1 & 49.9 \\
MAF~\cite{he2019multi} & 73.4 & 55.7 & 46.4 & 36.8 & 28.9 & 60.8 & 50.3 \\
Strong-Weak~\cite{saito2019strong} & \textbf{82.3} & 55.9 & 46.5 & 32.7 & \textbf{35.5} & 66.7 & 53.3 \\
\hline
\hline
ATF(ours) & 78.8 & \textbf{59.9} & 47.9 & \textbf{41.0} & 34.8 & \textbf{66.9} & \textbf{54.9} \\
\hline
\end{tabular}
\end{center}
\end{table}

\textbf{Transfer from Pascal VOC to Watercolor}. The Watercolor~\cite{inoue2018cross} dataset contains 6 categories which are the same as the Pascal VOC~\cite{everingham2010pascal}. In this experiment, the source only trained Faster-RCNN, DAF~\cite{chen2018domain}, WST-BSR~\cite{kim2019self}, MAF~\cite{he2019multi} and Strong-Weak~\cite{saito2019strong} are introduced for comparison. The results are shown in Table~\ref{table:water}, from which we can observe that our proposed ATF achieves the best performance among all compared models and the advantage is further proved.

\subsection{Cross-domain Detection from Synthetic to Real}

Domain adaption from the synthetic scene (auxiliary data) to the real scene is an important application scenario for domain adaptive object detection. If the domain adaption from synthetic data to real scene data is effective, the burden for the annotating the real images will be eased. In this section, we introduce the SIM10k~\cite{johnson2016driving} as the synthetic scene and the Cityscapes~\cite{sakaridis2018semantic} as the real scene. The source only trained Faster-RCNN, DAF~\cite{chen2018domain}, MAF~\cite{he2019multi}, Strong-Weak~\cite{saito2019strong}, and SCL~\cite{shen2019scl} are introduced for comparisons. The AP of the car is computed for the evaluation of the experiment which is presented in Table~\ref{table:sim10k}. We observe that our model outperforms all the compared models. The superiority of the proposed ATF is further demonstrated for the cross-domain object detection.
\begin{table}[h]
\begin{center}
\caption{The results of domain adaptive object detection on SIM10k and Cityscapes.}
\label{table:sim10k}
\begin{tabular}{|p{1.5cm}|p{1.5cm}|p{1.5cm}|p{1.5cm}|p{1.5cm}|p{1.5cm}|p{1.5cm}|}
\hline
Methods & F-RCNN & DAF~\cite{chen2018domain} & MAF~\cite{he2019multi} & S-W~\cite{saito2019strong} & SCL~\cite{shen2019scl} & ATF(ours) \\
\hline
\hline
AP(\%) & 34.6 & 38.9 & 41.1 & 40.1 & 42.6 & \textbf{42.8} \\
\hline
\end{tabular}
\end{center}
\end{table}
\subsection{Analysis and Discussion}

In this section, we will implement some experiments to analyze our ATF model with four distinct aspects, including parameter sensitivity, accuracy of classifiers, IOU v.s. detection performance and visualization.

\begin{figure}[t]
\centering
\subfigure[Accuracy of Classifiers]
{\includegraphics[width=0.32\linewidth]{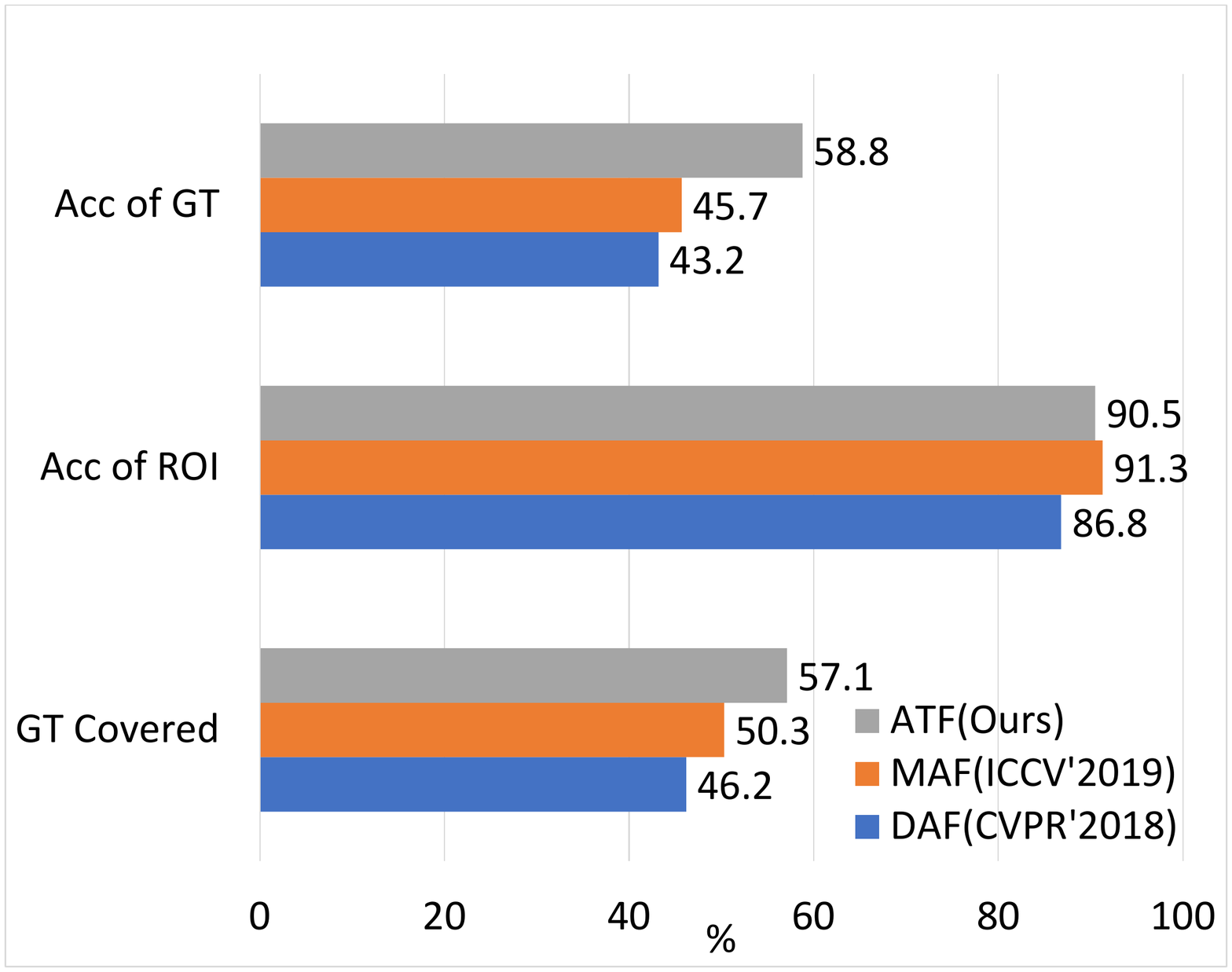}}
\subfigure[Performance v.s IOUs]
{\includegraphics[width=0.32\linewidth]{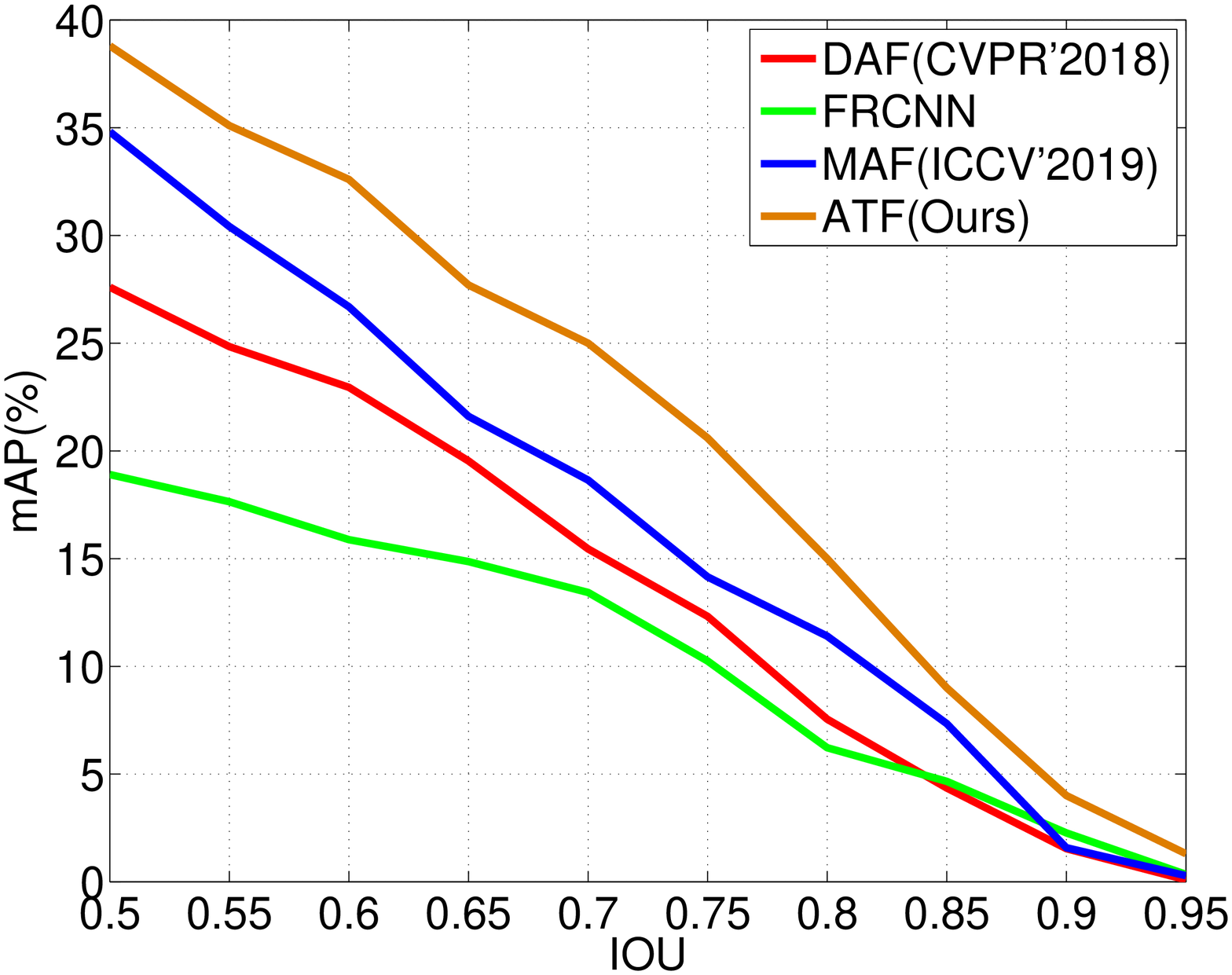}}
\subfigure[Performance v.s Epoch]
{\includegraphics[width=0.32\linewidth]{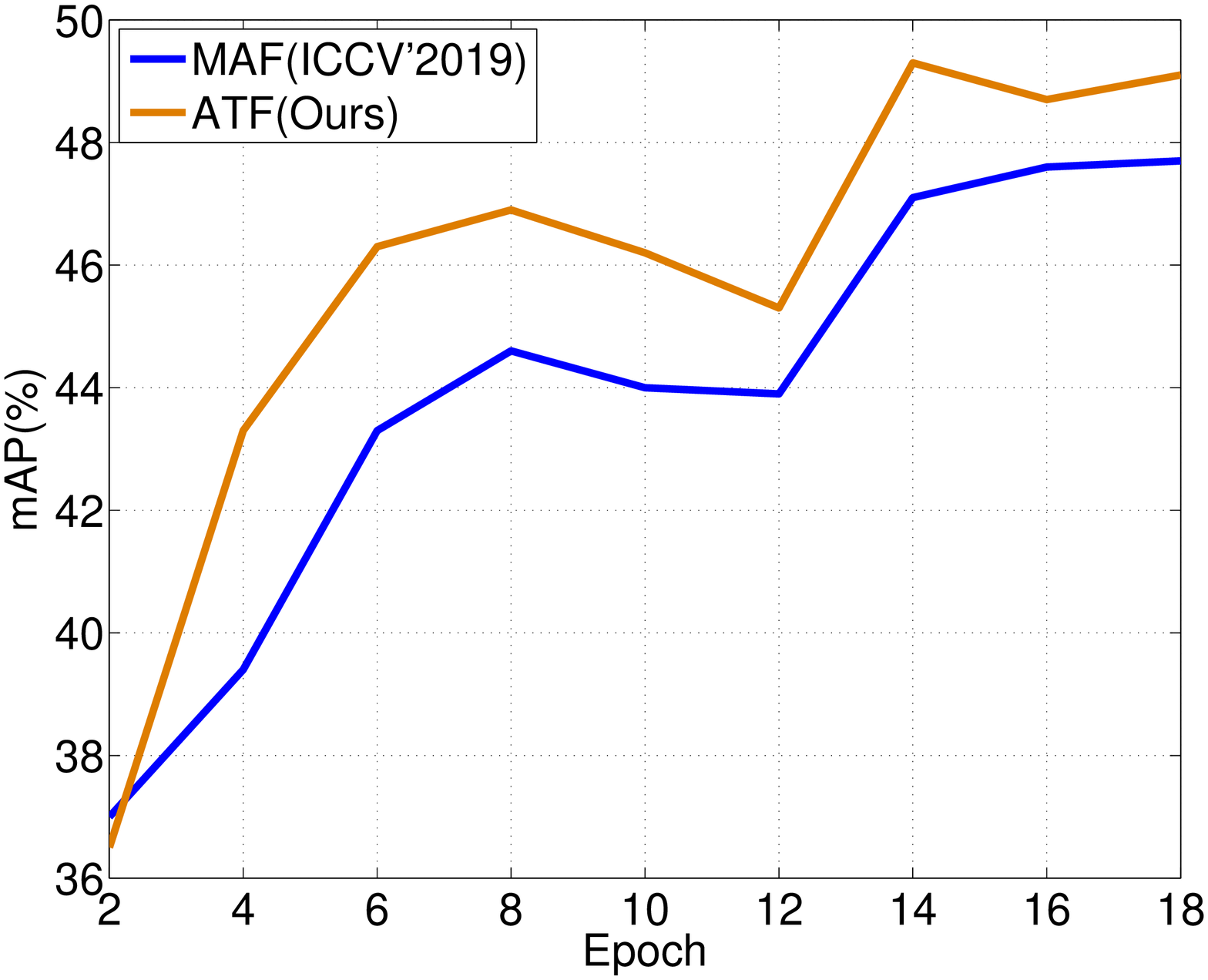}}
\caption{Analysis of our model. (a) We test the accuracy of the classifier in the detector. \textbf{Acc of GT}: We crop the ground truth of the image and use the classifier on R-CNN to classify them. \textbf{Acc of ROI}: The accuracy of RCNN's classifier with the ROIs generated by RPN. \textbf{GT Covered}: The proportion of ground truth seen by RPN. (b) The performance change with different IOUs. Better viewed in color version. (c) The performance of source domain with different epoches.}
\label{fig:exp}
\end{figure}

\textbf{Parameter sensitivity on $\alpha$}. In this part, we show the sensitivity of parameter $\alpha$ in Eq.(\ref{eq3}). $\alpha$ controls the power of domain adaption. We conduct the cross-domain experiments from Cityscapes to Foggy Cityscapes. The sensitivity of $\alpha$ is shown in Table~\ref{table:sen}. When $\alpha=0.7$, our model achieves the best performance.

\begin{table}[h]
\begin{center}
\caption{Parameter Sensitivity on $\alpha$.}
\label{table:sen}
\begin{tabular}{|p{1.5cm}|p{1cm}|p{1cm}|p{1cm}|p{1cm}|p{1cm}|p{1cm}|p{1cm}|}
\hline
 $\alpha$ & 0.3 & 0.5 & 0.7 & 0.9 & 1.1 & 1.3\\
\hline
\hline
mAP(\%) & 36.7 & 37.5 & 38.7 & 38.7 & 38.4 & 38.0 \\
\hline
\end{tabular}
\end{center}
\end{table}

\textbf{The accuracy of classifiers}. Our model enhances the training of the detector with the ancillary target feature. In this part, we analyze the classifier in the detector with different models. First, we static the number of ground truth boxes covered by the ROIs. If the IOU between the ROI and ground truth is higher than 0.5, we think the corresponding ROI is predicted by the RPN (region proposal network). Second, we compute the accuracy of the RCNN classifier with the ROIs from the RPN. Last, we crop the ground truth of the testing set and use the RCNN classifier to predict their label and get the accuracy. The DAF~\cite{chen2018domain}, MAF~\cite{he2019multi} and our model are tested with Cityscapes and Foggy Cityscapes datasets. The results are shown in Fig.~\ref{fig:exp}(a).
We observe that the RPN of our model finds 57.1\% of the ground truth, which is the best result of all compared models. The RCNN classifiers from all three tested models achieve above 90\% accuracy when classifying the generated ROIs. However, when the cropped ground truth samples are fed into the model, the accuracy of the RCNN classifier sharply drops. The ground truth samples missed by RPN are also misjudged by the RCNN in the experiments. Therefore, we experimentally find that it is very important to improve the recall of RPN in the cross-domain object detection task. In our ATF, benefited by the asymmetric structure which enhances the structural discrimination for the detector and the new decision boundary contributed by ancillary target features, our model achieves a preferable recall in RPN. This motivation is experimentally proved.

\textbf{Detection performance w.r.t. different IOUs}. The IOU threshold is an important parameter in the test phase. In the previous experiments, the IOU threshold is set as 0.5 by default. In this part, we test our model with different IOUs. The source only Faster-RCNN, DAF~\cite{chen2018domain}, MAF~\cite{he2019multi}, and our ATF are implemented for comparison. We conduct the experiments on the Cityscapes~\cite{cordts2016cityscapes} and Foggy Cityscapes~\cite{sakaridis2018semantic} datasets. The results are shown in Fig.~\ref{fig:exp}(b), where the IOU threshold is increased from 0.5 to 0.95. The performance drops as the IOU threshold increases in the experiment. Our model achieves the best performance on all tested IOU thresholds.

\textbf{Source performance w.r.t. epoches in monitoring source risk}. In this part, we monitor the training process of adaption from Cityscapes~\cite{cordts2016cityscapes} to Foggy Cityscapes~\cite{sakaridis2018semantic}. The mAP(\%) on the test set of Cityscapes is shown in Fig.~\ref{fig:exp}(c). Our ATF achieves higher mAP compared to the parameter shared MAF~\cite{he2019multi} during the training phase. Benefited by the asymmetric structure, our ATF can well prevent the collapse of the source domain and preserve the structural discrimination of source features. This experiment fully proves that parameter sharing of network will deteriorate the empirical source risk $\epsilon_S(h)$, which then leads to high target risk $\epsilon_T(h)$. Thus, a safer adaption of our ATF than the parameter shared MAF is verified.

\begin{figure}[t]
\centering
{\includegraphics[width=1.0\linewidth]{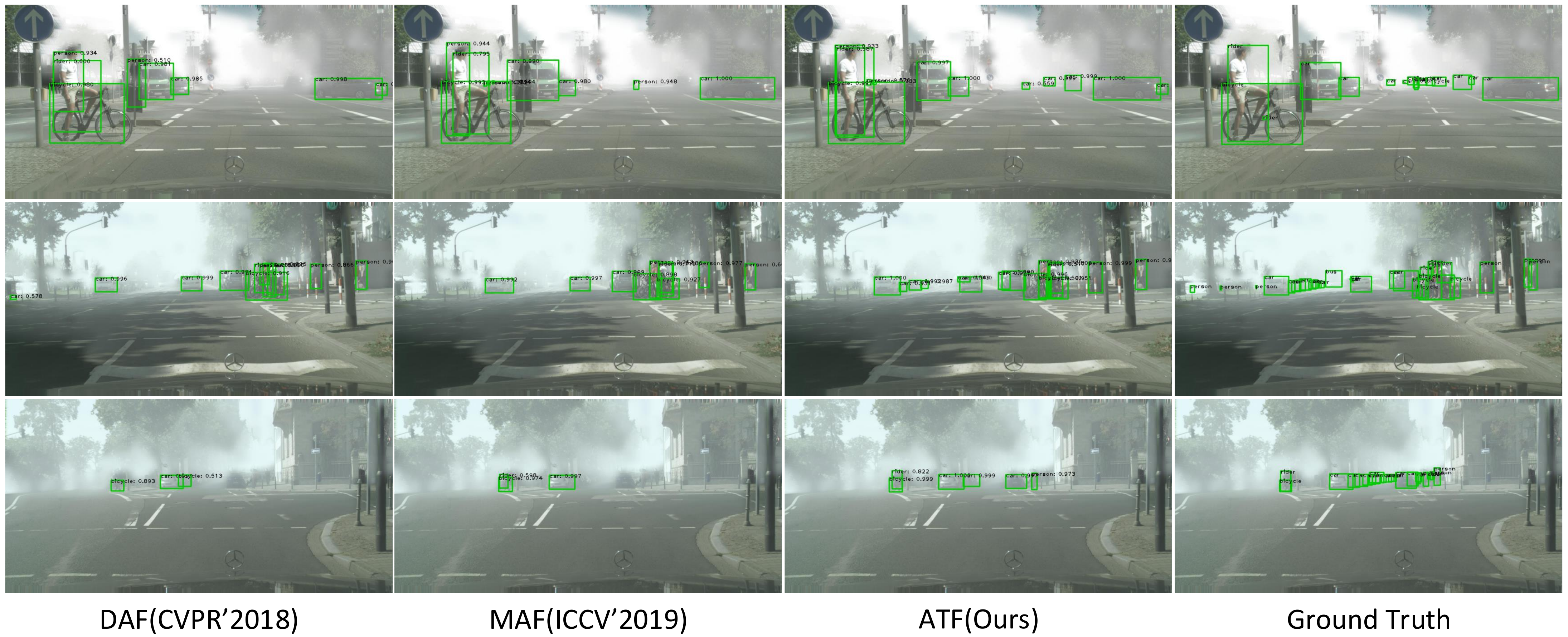}}
\caption{The visualization results on the target domain (Foggy Cityscapes~\cite{sakaridis2018semantic}).}
\label{fig:vis}
\end{figure}

\textbf{Visualization of domain adaptive detection}. Fig.~\ref{fig:vis} shows some qualitative object detection results of several models on the Foggy Cityscapes dataset~\cite{sakaridis2018semantic}, i.e. target domain. The state-of-the-art models, DAF~\cite{chen2018domain} and MAF~\cite{he2019multi}, are also presented. We can clearly observe that our ATF shows the best domain adaptive detection results and better matches the ground-truth.
\section{Conclusions}
In this paper, we propose an asymmetric tri-way network (ATF) to address the out-of-control problem of parameter shared Siamese transfer network for unsupervised domain adaptive object detection. In ATF, an independent network, i.e. the ancillary net, supervised by source labels, is proposed without parameter sharing. Our model has two contributions: 1) Since the domain disparity is hard to be eliminated in parameter shared siamese network, we propose the asymmetric structure to enhance the training of the detector. The asymmetry can well alleviate the labeling unfairness between source and target. 2) The proposed ancillary net enables the structural discrimination preservation of source feature distribution, which to a large extent promotes the feature reliability of the target domain. Our model is easy to be implemented end-to-end for training the chief net and ancillary net. We conduct extensive experiments on a number of benchmark datasets and state-of-the-art results are obtained by our ATF.

%
%
\bibliographystyle{splncs04}
\bibliography{egbib1}

\end{document}